\documentclass[runningheads]{llncs}
\usepackage{graphicx}
\usepackage{amsmath, amssymb}
\usepackage{subcaption}
\usepackage{url}
\usepackage{xcolor}

\begin{document}
\title{Knowledge Acquisition and Completion for Long-Term Human-Robot Interactions using Knowledge Graph Embedding}
\titlerunning{Knowledge Acquisition and Completion for Long-Term HRI}
%
\author{
Ermanno Bartoli\textsuperscript{\textsection} \and Francesco Argenziano\textsuperscript{\textsection}
\and \\ Vincenzo Suriani 
 \orcidID{0000-0003-1199-8358}  \and
 Daniele Nardi
 \orcidID{0000-0001-6606-200X}
}
\authorrunning{Bartoli et al.}
%
\institute{Dept. of Computer, Control, and Management Engineering\\ Sapienza University of Rome, Rome (Italy),
    \email{\{lastname\}@diag.uniroma1.it.}
}
\maketitle              
\begingroup\renewcommand\thefootnote{\textsection}
\footnotetext{These two authors contributed equally.}
\endgroup
\begin{abstract}
In Human-Robot Interaction (HRI) systems, a challenging task is sharing the representation of the operational environment, fusing symbolic knowledge and perceptions, between users and robots. 
With the existing HRI pipelines, users can teach the robots some concepts to increase their knowledge base. Unfortunately, the data coming from the users are usually not enough dense for building a consistent representation. Furthermore, the existing approaches are not able to incrementally build up their knowledge base, which is very important when robots have to deal with dynamic contexts. To this end, we propose an architecture to gather data from users and environments in long-runs of continual learning. We adopt Knowledge Graph Embedding techniques to generalize the acquired information with the goal of incrementally extending the robot's inner representation of the environment. 
We evaluate the performance of the overall continual learning architecture by measuring the capabilities of the robot of learning entities and relations coming from unknown contexts through a series of incremental learning sessions.
\end{abstract}
\section{Introduction}

\begin{figure}[h]
    \centering
    \includegraphics[width=1.0\textwidth]{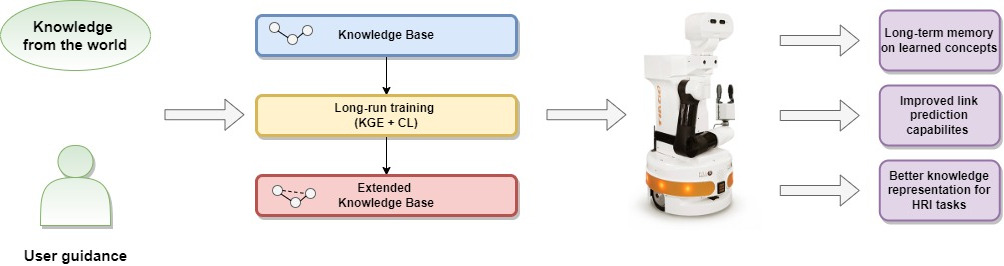}
    \caption{Complete architecture of the system: from the interaction with the user to the deployment of learned knowledge and capabilities after long-run training.}
    \label{fig:complete_arch}
\end{figure}

In the last years, robots started leaving laboratories to enter our daily environments where they are asked to autonomously operate, often sharing the working area with humans. To be effective in this goal, representing and storing information in a suitable way is fundamental regardless of the specific robotic applications. In particular, this problem acquires more relevance when designing Human-Robot Interaction (HRI) systems, since there is the intrinsic need to make the human and the robot participants interact with each other. In order to make this interaction successful, the robot and the human not only must be able to communicate and understand each other, but also they should have a mutual understanding of the world they both operate in. Therefore, a shared semantic of the environment is needed in order to make the interaction successful. In many HRI applications, this knowledge (that is the building block on which the whole system is built) is often embedded in the agent's behaviour and changes from one episode to another. A way to improve it can be through a generalization of the knowledge that is transferred and acquired by the robot. In fact, usually, it is very domain-dependent 
for the specific application of the system. 
%

In this paper, we propose a novel architecture for acquiring knowledge from sparse data acquisition from environments. The acquired knowledge is represented and organized 
to improve the completeness of the previous knowledge base of the robot. This process leads to the creation of a resulting more extensive knowledge base that is built up incrementally. The nature of the architecture is meant to be robust to any change in the context so that it can be suitable in several HRI applications, even if very different from each other. A major advantage of the proposed approach is that, differently from previous HRI systems, it is not necessary to modify the software architecture when the context of the interaction changes, but it is only needed to start a new learning session that shapes the existing learning skills of the robot. The acquisition of the data is human-driven, and the human who collaborates with the robot is not required to know anything about the software of the agent, nor how the knowledge is represented, but the user just needs to share his knowledge of the world with the robot. This process needs to take into account some aspects. First of all, this kind of interaction is not defined over a short period of time, long-runs are necessary to achieve good results. However, long-runs are not that common in the HRI field, since the interactions between humans and robots happen quite fast, and therefore this problem must be treated. Moreover, because of these long-runs, the robot will face information that needs to be stored and effectively processed, without forgetting acquired knowledge as the run goes on. To solve these problems, the methodology we propose relies on Continual Learning (CL) and Knowledge Graph Embeddings (KGEs): the former is used to deal with the catastrophic forgetting phenomenon during incremental knowledge acquisition sessions, while the latter is used to efficiently use the information, stored in a Knowledge Graph (KG) database, to perform the knowledge completion. In the end, the knowledge of the system spans from grounded facts about the environment to more general concepts on which the system can make predictions. This knowledge allows for several reasoning processes, based on the kind of query that the human operator may ask: if the query is very specific (namely the human asks for a particular object in a particular location), the robot can answer by exploiting its \textit{experience}, that is what it has detected in the past explorations; for more general queries (namely, general objects or concepts), the robot can answer by making predictions depending on what it has learned, so by using an \textit{ontological scheme} of the environment that it has slowly built in the past days.

\begin{figure*}[t!]
    \centering
    \includegraphics[width=1.0\linewidth]{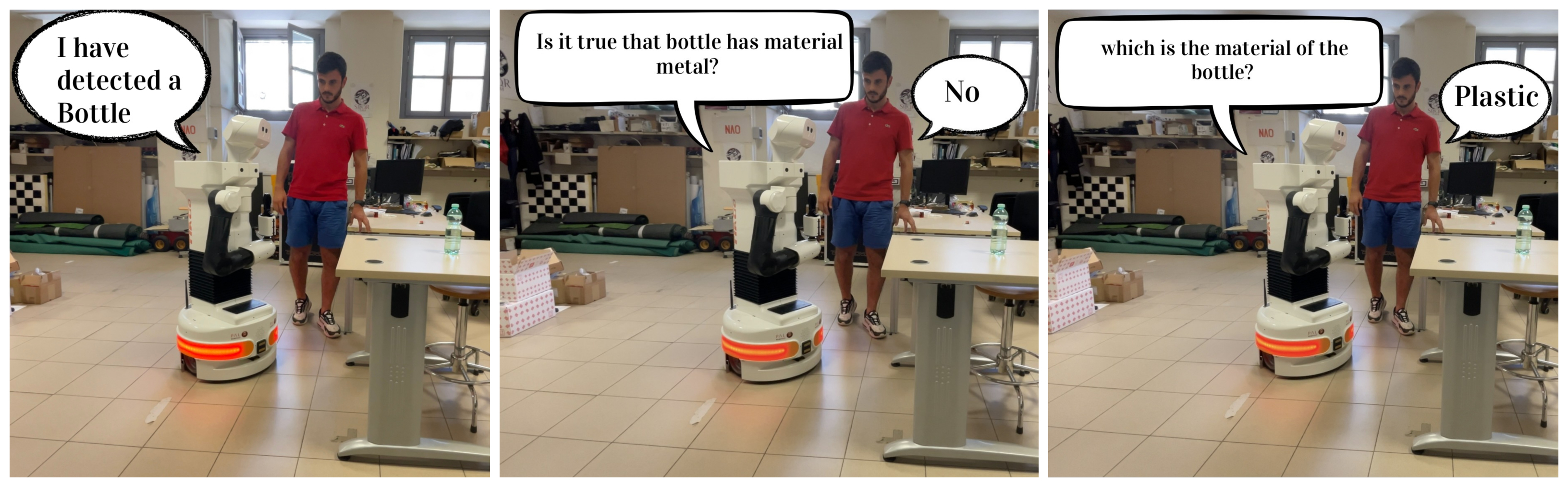}
    \caption{Interaction with the robot before the long-run training and knowledge acquisition. The robot still has difficulties in carrying on a correct interaction.}
    \label{fig:prev}
\end{figure*}

\begin{figure*}[t!]
    \centering
    \includegraphics[width=1.0\linewidth]{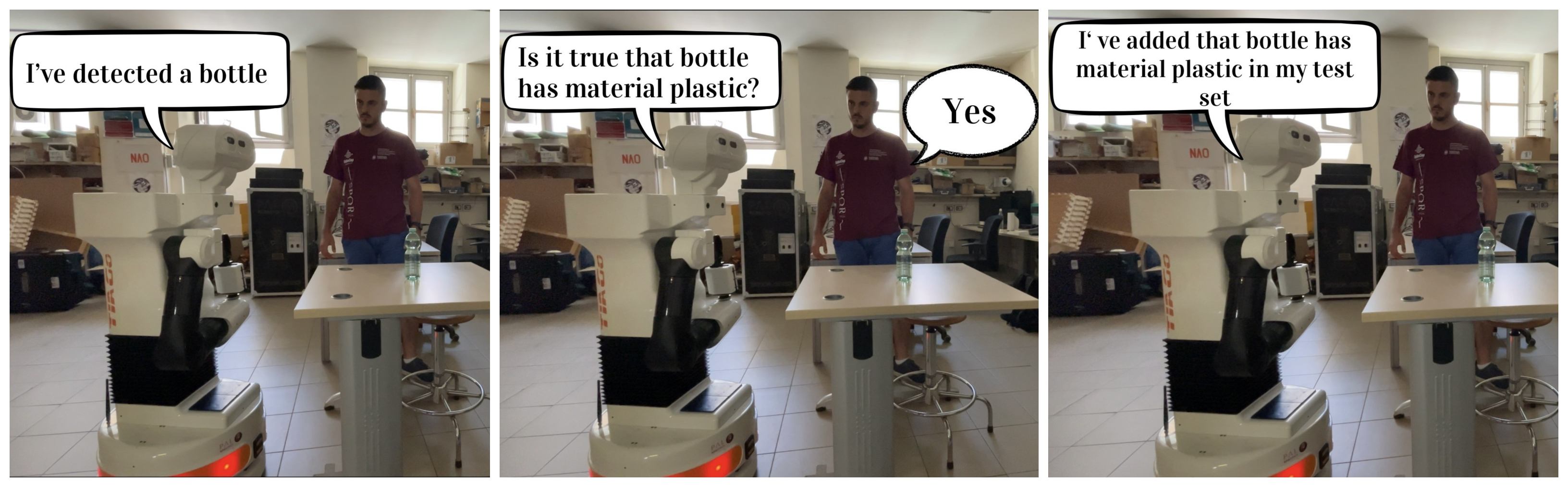}
    \caption{Interaction with the robot after the long-run training and knowledge acquisition. The robot has improved its capabilities, can correctly carry on the interaction, and exploits it to learn new relations.}
    \label{fig:after}
\end{figure*}

\section{Related Work}
In order to have robots working and acting in human-shaped environments, semantic mapping approaches have been studied, aiming at constructing a common representation of the world between robots and humans \cite{pronobis2011semantic}. To this end, there was a growing need of representing the knowledge of the robot with appropriate techniques in order to allow for faster and more precise reasoning about the environment the agents lived in. One particular way of knowledge representation that is demonstrated to be very effective is through triples \cite{pronobis2012large}, in which objects of the worlds are linked together by some sort of relation. This way of memorizing facts enabled the usage of a particular kind of data structure, the Knowledge Graphs (KGs) \cite{ji2021survey}, in which is it possible to represent collections of triples as directed graphs. In those graphs, objects and entities are represented as nodes, and relations between entities are represented as directed edges. This representation allows for better data integration, unification, and reuse of information since it is also easier to represent ontological structures by the use of them. However, one of the biggest problems of KGs is that they do not scale well with size: the bigger the graph, the harder is to navigate through it and the harder is to make any sort of inference from it. For this reason, instead of working directly with KGs, through the years techniques of Knowledge Graph Embeddings (KGEs) \cite{wang2017knowledge} have been developed, in which KGs are transformed into lower-dimensional representation in order to reduce the number of parameters of the system while also preserving the information of the graph. Another problem in representing information with KGs is that when knowledge comes from multiple sources, there is often the possibility of incorporating contradictory pieces of information that will eventually compromise the quality of the system (in particular during the training of the embedding). For this reason, it is important to introduce in the process of knowledge acquisition some sort of validation procedure, and this validation can be done by interacting with humans. In recent years, the human participant in the interaction has acquired a bigger and bigger role in the robot's acquisition of knowledge from the world \cite{gemignani2016living} \cite{randelli2013knowledge}, and this is because through the filtering process of a human we are able to transfer to the robot only useful information, that can significantly improve further reasoning processes down the interaction pipeline. Although the human can get rid of useless information, a human-drive acquisition of knowledge needs much time to be robust and efficient, because the data that the robot acquires through the human can be sparse and not cohesive. For that purpose, the development of systems capable to handle long-runs of one single experiment has become more popular \cite{long_run1}. This kind of experiment allows the robot to build up robust and dense knowledge. An interesting way to build up the robot's knowledge is doing it incrementally through human-robot interaction.
Such a class of problems has been addressed in applications focused on learning without forgetting \cite{learning_without_forgetting}. These approaches typically operate in a task-based sequential learning setup. This formulation, which is rarely encountered in practical applications under this assumption, has been also studied in a task-free scenario \cite{aljundi2019task}.

\section{Methodology}
The proposed approach aims at making the robot able to address the multi-relational embedding problem while incrementally building up the robot's knowledge base in a unique long-run. The goal mentioned can be subdivided into three subtasks which are addressed at the same time: acquiring data in collaboration with the human, incorporating the acquired data in the infrastructure designed for semantic mapping, improving the accuracy of the robot's predictions by training the model on the new data.
\begin{figure}[h]
    \centering
    \includegraphics[width=0.6\textwidth]{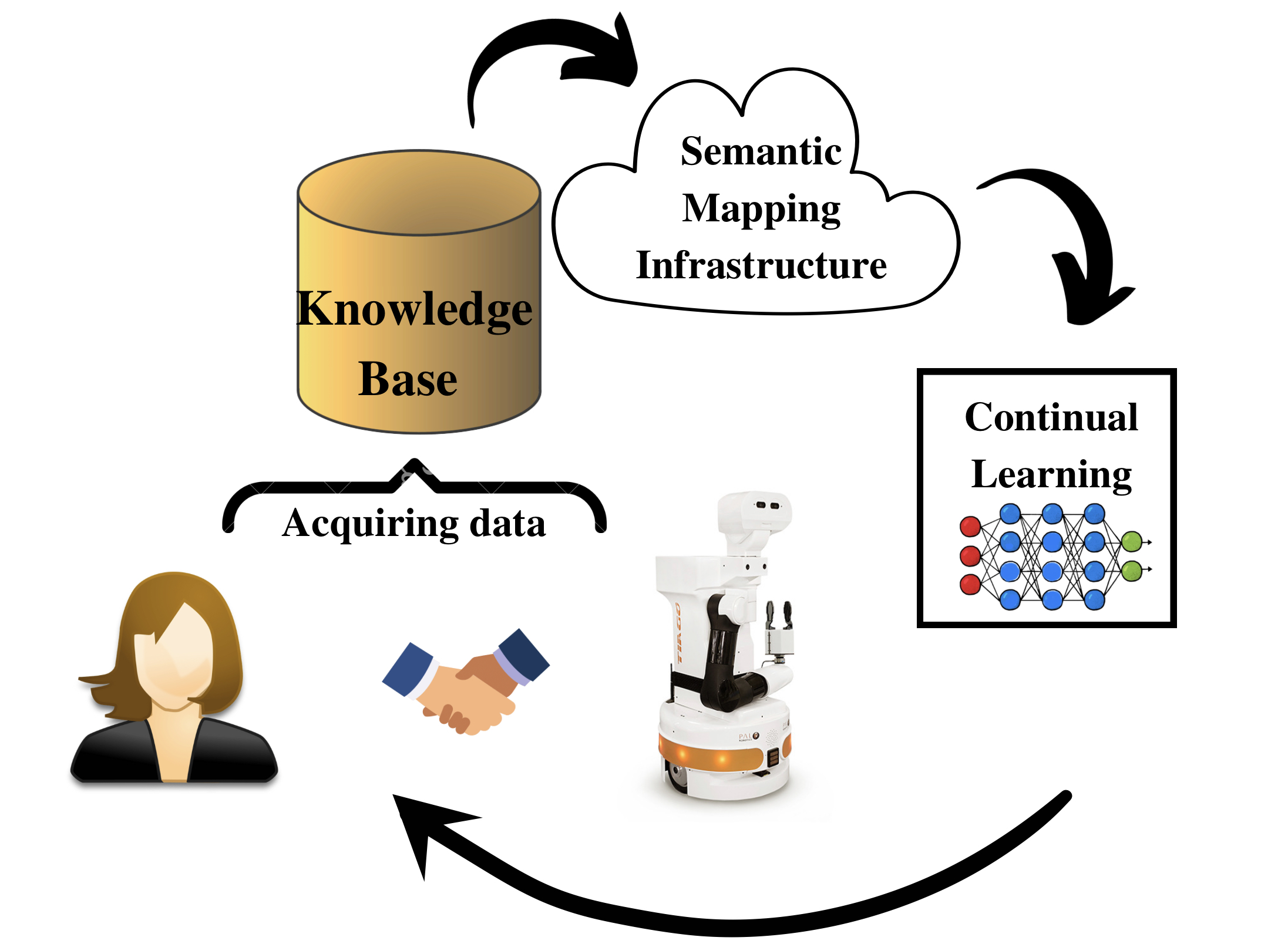}
    \caption{The final task is the composition of three sub-tasks.}
    \label{fig:sub_problems}
\end{figure}

\subsection{Acquiring and Extending the Knowledge Base}
\begin{figure}[t]
    \centering
    \includegraphics[width=\textwidth]{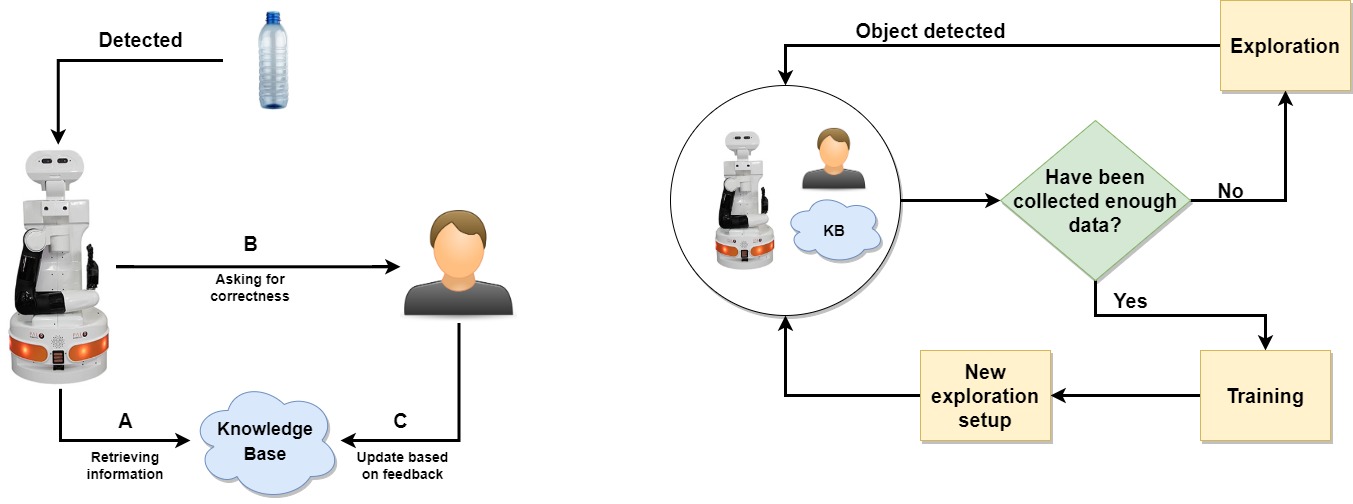}
    \caption{On the left, the process of acquiring meaningful information, composed by 3 phases: retrieving information (A), asking for correctness (B), and updating based on feedback (C). On the right, the workflow for a long-run execution.}
    \label{fig:block_interaction}
\end{figure}

To properly build a knowledge base (KB) for the purpose of this work, we chose to have a basic predicate represented by a \textit{triple}, $(h, r, t)$,
where $h$ is the head entity, $t$ is the tail entity, and $r$ is the relation that connects the head to the tail. A set of those triples can be suitable for Continual Learning on Knowledge Graph Embedding. In fact, a dataset of triples can be easily split into learning sessions, each of them comprising a portion of the data. This can be used to simulate the fact that data are not all available at once, so in the training session $n$, only the $n-th$ portion of the dataset is given to the model, and it trains itself only on those data. This procedure is valid, but it is assumed that even if the dataset is not given to the model entirely, it must be known in advance in order to be able to divide it. This is a huge constraint when dealing with real robots and real environments for two main reasons. The first is that, when the robot is put into an environment 
, the number and the type of the object in the environment are unknown. This means that the number of predicates that the robot collects when evolving in the environment, so the number of entities and relations of the robot's knowledge base, can vary. 
The second reason is that also the number of tasks can vary. In fact, 
when the robot detects an unknown object, the system has to take care of a new entity but also a new task. The architecture will assign an embedding to the new entity and the next training will include also such an entity. 
From a conceptual point of view, the interaction between the robot and the human that cooperate in order to enlarge the knowledge base is shown in Fig. \ref{fig:block_interaction}, on the left.
\noindent
In the context of Interactive Task Learning (ITL) \cite{laird2017interactive}, the setup of our experiments aims at developing agents and systems that are focused on generality across different tasks. In fact, the main focus is the ability of the system to abstract concepts from different domains on a more general level. Our work, which exploits embedding algorithms on the triples of a KG, adopts these principles.

The knowledge acquisition procedure consists of three different phases, that are chronologically consecutive. First, the objects detected using the YOLO Neural Network \cite{redmon2018yolov3} come into the robot as simple labels, and the \textbf{phase A} starts. The robot queries its KB in order to retrieve the semantic meaning of the object detected. The semantic meaning could be also inaccurate: in fact, the more that entity appears in the KB, the more the embedding of that entity will be precise and, the predictions on that entity, more accurate. If there are not enough data that grant an accurate embedding of the entity, the predictions will be incorrect. 
The predictions are represented by the predicates $(h,r,t)$ where the head entity is the detected object, the relation is chosen randomly among all the known relations, and the tail entity is the result of the prediction. After the generation of the predicates, the \textbf{phase B} starts. Here the robot asks the human for the correctness of the predicates by asking questions for each predicate. Communication is very important and it needs to be well-defined because misunderstanding could provoke incorrect answers that lead to the addition of wrong data to the KB. Since the data of the KB are not always human interpretable ("objInLoc" stands for "object is in Location"), according to the relation of the predicates, the question is generated so that it is human-understandable. As soon as the robot asks the question to the user, it waits for the user's answer, and \textbf{phase C} starts. In this phase, the user can answer positively or negatively. If the user answers positively, it means that the robot's prediction was correct, and the predicate $(h,r,t)$ is a true fact, so it can be added to the KB. If the prediction is judged as false, the robot asks the user for the correct tail of the predicate $(h,r,?)$, where $h$ and $r$ are the same head entity and relation as before. Once the user answers the robot with the correct tail entity, a new predicate is created, and it is added to the KB. In the end, both a correct prediction and an incorrect prediction lead to an addition of a true predicate in the KB. Moreover, when the robot adds the predicate to its KB, it provides an implicit consensus to the user. 
In this way, the user is able to know which predicate is being added to the knowledge base, and if there is an error, the user can recover from it.

\subsection{Knowledge Graph Embedding}
In order to predict new predicates, we adopted the Knowledge Graph Embedding (KGE) technique, which uses supervised learning models capable to learn vector representations of nodes and edges.
By definition, the objective of Knowledge Graph Embedding problem is to learn a continuous vector representation of a Knowledge Graph $\mathcal{G}$ which encodes vertices that represent entities $\mathcal{E}$ as a set of vectors $v_{\mathcal{E}} \in \mathbb{R}^{|\mathcal{E}| \times d_{\mathcal{E}}}$, where $d_{\mathcal{E}}$ is the dimension of the vector of entities $\mathcal{E}$, and as a set of edges which represent relations $\mathcal{R}$ as mappings between vectors $W_{\mathcal{R}} \in \mathbb{R}^{|\mathcal{R}| \times d_{\mathcal{R}}}$, where  $d_{\mathcal{R}}$ is the dimension of the vector of relations. The knowledge graph $\mathcal{G}$ is composed by triples $(h,r,t)$, where $h,t \in \mathcal{E}$ are the head and tail of the relations, while $r \in \mathcal{R}$ is the relation itself. One example of such a triple is (\emph{bottle, hasMaterial, plastic}). In literature, there are numerous ways of embedding the knowledge in a knowledge graph: \textit{transitional models, rotational models, gaussian models}, and many others. However, independently on what is the class of methods that are used, the embedding is learned by minimizing the loss $\mathcal{L}$ computed on a scoring function $f(h,r,t)$ over the set of triples in the knowledge graph, and over the set of negative triples that are generated by negative sampling over the same graph. For this research, the embedding model we used is ANALOGY 
that represents a relation as a matrix. This model can cope with asymmetrical relations and imposes the structure of the matrix to be a diagonal-block matrix, to minimize the number of parameters that need to be stored by the system.

\subsubsection{ANALOGY}
In the field of KGEs, there are many numerous ways of representing the relations into lower dimensional spaces. Usually, these techniques are grouped in families of models that describe the general principle that makes the embedding of the information possible. For instance, translational models (like TransE \cite{bordes2013translating}) represent relationships as translations in the embedding space, while Gaussian embeddings model also takes the uncertainty of the information contained in a KG. Despite these models being simpler than other models, they fail to correctly represent more complex kinds of relations (like symmetrical relations), and so more advanced models are needed. For this reason, we chose ANALOGY as our KGE model. ANALOGY is an improvement of the RESCAL \cite{nickel2011three} model that is a tensor factorization approach able to perform collective training on multi-relational data. In this approach, a triple $(h.r.t)$ is represented as an entry in a three-way tensor $\mathcal{X}$. A tensor entry $\mathcal{X}_{ijk}=1$ means that the triple composed by the i-th and the k-th entity as, respectively, head and tail, and the j-th relation is a true fact. Otherwise, unknown or non-existing facts have their entry set to 0. Each slice $\mathcal{X}_k$ of the tensor is then factorized as $\mathcal{X}_k\approx AR_kA^T$, where $A$ is a matrix that contains the latent-component representation of the entities, while instead $R_k$ is a matrix that models the interactions of the latent components, and both are computed by solving the minimization problem 
\begin{equation}
\label{min_prob}
    \min _{A, R_{k}} f\left(A, R_{k}\right)+g\left(A, R_{k}\right)
\end{equation}
where
\begin{equation}
\label{f}
    f\left(A, R_{k}\right)=\frac{1}{2}\left(\sum_{k}\left\|\mathcal{X}_{k}-A R_{k} A^{T}\right\|_{F}^{2}\right)
\end{equation}
and $g$ is a regularization term
\begin{equation}
\label{g}
    g\left(A, R_{k}\right)=\frac{1}{2} \lambda\left(\|A\|_{F}^{2}+\sum_{k}\left\|R_{k}\right\|_{F}^{2}\right)
\end{equation}
Starting from this approach, ANALOGY makes some important improvements: it constrains $R$ to be a diagonal matrix (like DistMult), and it introduces complex-valued embeddings to cope with asymmetric relations $X=EW\Bar{E}^T$ (like ComplEx does), but most importantly it imposes analogical structures among the representations by the means of a diagonal-block matrix (reducing the number of parameters needed by the model) by modifying the objective function as follows 
\begin{equation}
\label{dl}
    \begin{array}{cl}
\min _{v, W} & \mathbb{E}_{s, r, o, y \sim \mathcal{D}} \ell\left(\phi_{v, W}(s, r, o), y\right) \\
\text { s.t. } & W_{r} W_{r}^{\top}=W_{r}^{\top} W_{r} \forall r \in \mathcal{R} \\
& W_{r} W_{r^{\prime}}=W_{r^{\prime}} W_{r} \quad \forall r, r^{\prime} \in \mathcal{R}
\end{array}
\end{equation}
 
\subsection{Long-Run}
 
The process described is robust, because allows a robot that is put in a completely unknown environment, to incrementally build a robust knowledge of it. A completely unknown environment means that no entity or relation is present in the KB of the robot at the beginning. Moreover, one of the advantages of this approach is that some knowledge could be transferred to the robot. For example, it is possible to exploit existing knowledge graph databases to give some a-priori knowledge to the robot. In this way, the robot will learn to build up its KB much faster.
During this process, the KB of the robot evolves in the environment, acquiring information and communicating with the human. 
This approach is meant for designing a single long-run, instead of multiple short runs. 
Fig. \ref{fig:block_interaction}, on the right, shows the block scheme of such approach.

The circular block, depicting the robot and the user, wraps all the infrastructure responsible for enlarging the KB and communicating with the human, which is shown in Fig. \ref{fig:block_interaction}, on the left. The two blocks, i.e. \textit{exploration} and \textit{training}, are mutually exclusive. 
These 2 blocks are called whether or not a condition is verified. There are three different conditions that have been implemented. The first (shown in Fig. \ref{fig:block_interaction}, on the right) deals with the amount of data collected by the robot during the exploring phase. This kind of condition makes it possible that at each learning session the robot collects the same amount of data, so the dataset will always be balanced. The second condition deals with the battery level of the robot. With this condition, the robot is free to explore the environment until the battery goes under a certain threshold, so the robot comes back to its docking station and, while recharging, it performs a training session. The final condition only includes time. Two periods, namely \textit{day} and \textit{night}, are defined. In the first one, the robot is in exploration, while in the latter, the robot is in training.

\section{Results}
\begin{table*}[t]
\centering
\begin{tabular}{ |c||c|c|c|c|c|c| } 
\hline
 & sess\_0 & sess\_1 & sess\_2 & sess\_3 & sess\_4 & sess\_5 \\
\hline
\hline
classical\_context on ai2thor\_5 & - & - & - & - & - & 0.705 \\ 
classical\_context on ai2thor\_4 & - & - & - & - & 0.238 & 0.764 \\ 
classical\_context on ai2thor\_3 & - & - & - & 0.346 & 0.336 & 0.676 \\ 
classical\_context on ai2thor\_2 & - & - & 0.382 & 0.371 & 0.389 & 0.647 \\ 
classical\_context on ai2thor\_1 & - & 0.402 & 0.385 & 0.361 & 0.380 & 0.558 \\ 
classical\_context on ai2thor\_0 & 0.339 & 0.355 & 0.343 & 0.343 & 0.336 & 0.500 \\ 
\hline
\end{tabular}
\caption{\textbf{HITS@10} of ANALOGY with Standard settings}
\label{table:1}
\end{table*}

\begin{table*}[t]
\centering
\begin{tabular}{ |c||c|c|c|c|c|c| } 
\hline
 & sess\_0 & sess\_1 & sess\_2 & sess\_3 & sess\_4 & sess\_5 \\
\hline
\hline
classical\_context on ai2thor\_5 & - & - & - & - & - & 0.569 \\ 
classical\_context on ai2thor\_4 & - & - & - & - & 0.104 & 0.385 \\ 
classical\_context on ai2thor\_3 & - & - & - & 0.129 & 0.128 & 0.338 \\ 
classical\_context on ai2thor\_2 & - & - & 0.136 & 0.127 & 0.130 & 0.322 \\ 
classical\_context on ai2thor\_1 & - & 0.153 & 0.146 & 0.141 & 0.146 & 0.270 \\ 
classical\_context on ai2thor\_0 & 0.151 & 0.134 & 0.134 & 0.130 & 0.130 & 0.198 \\ 
\hline
\end{tabular}
\caption{\textbf{MRR} of ANALOGY with Standard settings}
\vspace{-1.5em}
\label{table:2}
\end{table*}
In the evaluation of the presented work, we would like to capture the capability of the robot to exploits its knowledge during the process of learning whatever the human teaches to it.

The learning procedure is built so as to recognize the entities in a certain environment, also to learn the relations between these entities, and predict them even when they are not explicitly mentioned by the human. The first thing that we want to prove is that models based on the standard learning process tend to forget what they have learned when new things to learn come. 
In order to prove this, we have simulated with the TIAGo robot a situation in which it learns from the human some information belonging to a certain context, and then it is asked to learn other information from a different context. From a technical point of view, this experiment consists of training the robot over 6 learning sessions, using a dataset structure that is inspired by AI2THOR \cite{kolve2017ai2}. In the first one, the dataset sess\_5\_ai2thor has been taken as input. Instead, for the subsequent 5 learning sessions, the dataset  $sess\_i\_ai2thor$ with $i \in \{0,1,2,3,4\}$ has been used.
In particular, the dataset sess\_5\_ai2thor has been created by the robot through the methodology described. Moreover, the model used for this experiment is \textbf{ANALOGY}, and it has been developed in "classical\_context" which means that it has not been made suitable for continual learning, but it is such as the standard model for KGEs.\\
The results of this experiment are showed in tables \ref{table:1} and \ref{table:2}. The two tables show the performances of the model in terms of \textbf{HITS@10} (Hits at 10) and  \textbf{MRR} (Mean Reciprocal Rank).
The 2 metrics \textbf{HITS@10} and \textbf{MRR} are defined as follows:
\begin{equation}
    \label{eq:mrr}
    \centering
    MRR=\frac{1}{|Q|} \sum_{i=1}^{|Q|} \frac{1}{\operatorname{rank}_{(s, p, o)_{i}}}
\end{equation}

\begin{equation}
    \label{eq:hitsn}
    \centering
    Hits@10=\sum_{i=1}^{|Q|} 1 \text { if } \operatorname{rank}_{(s, p, o)_{i}} \leq 10
\end{equation}\noindent
Each table must be read from the top to the bottom because the order is chronological. In each row, there is the performance of the model (trained on the subset $i$ of the dataset) with respect to the other subsets.
The first row of \ref{table:1} for instance, shows the HITS@10 of ANALOGY which has been trained on sess\_5\_ai2thor.
Since it has only been trained on that subset of the dataset, it has been evaluated only on sess\_5\_ai2thor.
The row "classical\_context on 2\_ai2thor", shows the HITS@10 of ANALOGY which has been trained on sess\_5\_ai2thor, sess\_4\_ai2thor, sess\_3\_ai2thor (\textbf{previously}), and sess\_2\_ai2thor (\textbf{currently}). It means that can be evaluated on the subset $sess\_i\_ai2thor$ where $i \in \{2,3,4,5\}$. 
The model comes across the catastrophic forgetting phenomenon because, the more it trains on subsets $sess\_i\_ai2thor$ where $i \in \{4,3,2,1,0\}$ which contain the same entities and relations, the less it is precise on HITS@10 on sess\_5\_ai2thor whose data are unseen for all the subsequent learning sessions.

\begin{figure*}[t]
    \centering
    \includegraphics[width=\textwidth]{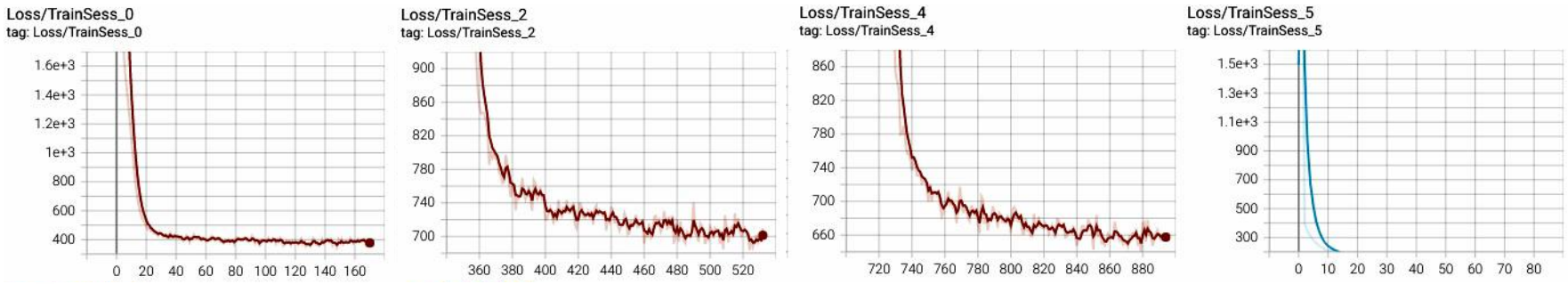}
    \caption{The \textit{Loss} during the learning sessions: 0, 2, 4, 5. (The last one, in blue, represents the training considering the last subset of the data acquired through the proposed methodology).
    This shows that the trend is constantly decreasing.}
    \vspace{-1.0em}
    \label{fig:graph1}
\end{figure*}
\begin{figure*}[t]
    \centering
    \includegraphics[width=\textwidth]{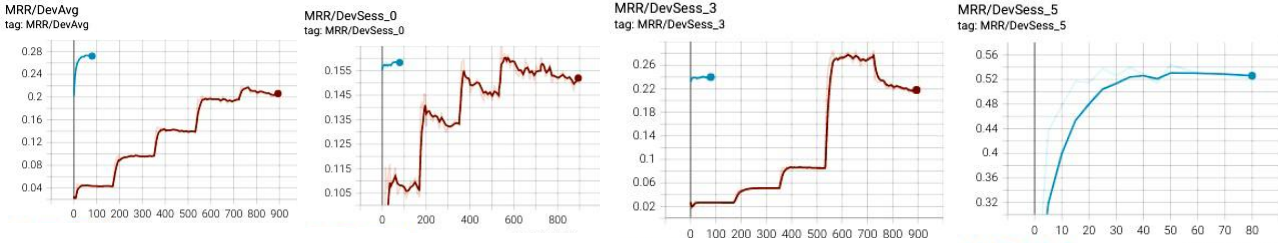}
    \caption{The \textit{MRR} during the learning sessions: 0, 3, 5 (the last one, in blue, represents the training considering the last subset of the data acquired through the proposed methodology).
    The graph on the left compare the MRR of last learning session with the average MRR among all the previous learning sessions.}
    \vspace{-1.5em}
    \label{fig:graph2}
\end{figure*}
\begin{figure*}[t]
    \centering
    \includegraphics[width=\textwidth]{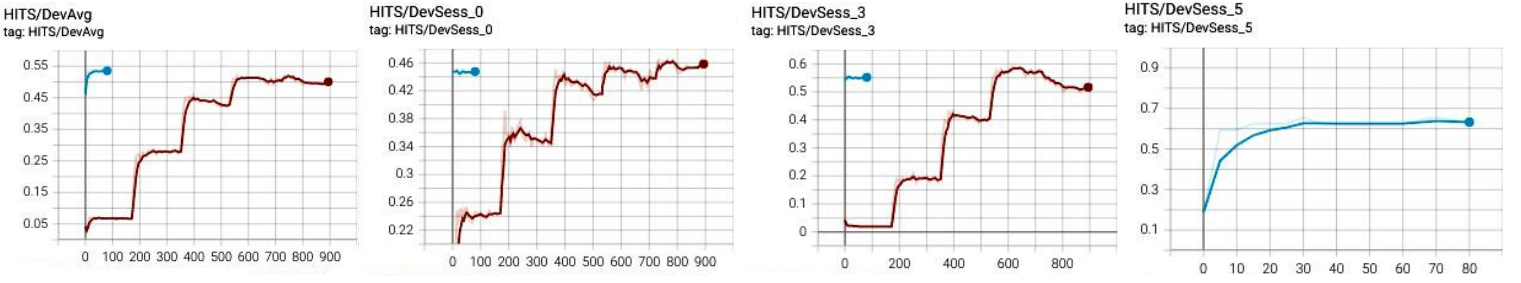}
    \caption{The HITS@10 function during the learning sessions: 0, 3, 5 (the last one, depicted in blue, represents the training considering the last subset of the data acquired through the proposed methodology).
    The graph on the left compares the HITS@10 of the last learning session with the average MRR among all the previous learning sessions.}
    \vspace{-1.5em}
    \label{fig:graph3}
\end{figure*}

For the next experiment, the model ANALOGY is considered only with continuous\_context, because it proves efficient for the problem of catastrophic forgetting. The same dataset considered previously has been used, i.e. $sess\_5\_ai2thor$ with $i \in \{0,1,2,3,4,5\}$, where the partitions $i \in \{0,1,2,3,4\}$ are composed by the same types of entities, while the partition $i = {5}$ consists mostly of new entities.\\
In Fig. \ref{fig:graph1} are shown 5 different graphs, representing the trend of the loss function for each learning session. By only looking at these graphs, there are some elements that are very important. First, the trend of the loss is always decreasing. The most decreasing shape is reached in the first forty epochs of  each learning session. Since in each learning session there is a limited amount of data, after some epochs the trend is quite stable, and the model is no longer improving. Here comes the "early stopping", which is set with a patience=50, that stops the training for that learning session and starts the next learning session. Although the entities are almost the same in each learning session, the predicates are different, and for this reason, at the beginning of each learning session the loss is pretty high, but then it decreases. The overall trend of the loss decreases learning session by learning session.\\
The loss function is an important metric for checking if the model is learning or not, but is not significant if considered alone, in fact Fig. \ref{fig:graph2} and Fig. \ref{fig:graph3} show the graphs of the two metrics considered for the evaluation of the models, which are \textit{MRR} and \textit{HITS@10}.
The increasing learning skills are confirmed by the graphs of MRR and HITS. The model, in fact, is not only evaluated on the $nth$ portion of the dataset given in input for training but all portions of the data are considered in the evaluation. Hence, if good performances were expected when evaluating the current portion of the dataset (see MRR/DevSess\_5 in \ref{fig:graph2} and HITS/DevSess\_5 in \ref{fig:graph3}), it was not sure that it was also for the previous ones. The results showed a remarkable ability to not forget what is learned, and it is visible in MRR/DevSess\_$i$ with $i \in \{0,1,2,3,4\}$ and HITS/DevSess\_$i$ with $i \in \{0,1,2,3,4\}$. In these graphs, the performance of the last learning session is marked with the color blue. Both for MRR and for HITS the performances of the last learning session (represented in blue color) are not worse than the performance of the model at the previous learning session (depicted in red).

Finally, when evaluating performances, it might be worth considering also if they are affected by \emph{performative effects} \cite{perdomo2020performative}. These phenomena have always been present in several fields like statistical decision theory and causal reasoning, but in the last years, they have been brought to attention also in the deep learning field. They can occur when predictions may influence the outcomes they are trying to predict, causing a distribution shift of the obtained results. It has been observed that these effects are reduced if multiple re-training procedures are performed. In the present work, we proposed a re-training procedure at the end of each learning session. This operation would reduce such distribution shifts.

A video representing a key result of this work can be found in the following link: \url{https://www.youtube.com/watch?v=vQbyn7hs8_4}. It shows, through some snapshots of the video, the process of enlarging the knowledge base of the robot, thanks to the interaction with the human. With this procedure, entities that were first unknown, become part of the knowledge of the robot.

\section{Conclusions and Future Directions}
In this work, we show (as in Figg. \ref{fig:prev} and \ref{fig:after}) the ability of the robot to learn from unknown environments, relying on the answers of the human. Thanks to the proposed architecture, the robot uses Knowledge Graph Embedding techniques to generalize the acquired information with the goal of incrementally extending its inner representation of the environment. We evaluate the performance of the overall architecture by measuring the capabilities of the robot of learning entities and relations coming from unknown contexts through a series of incremental learning sessions, demonstrating the ability of the presented architecture to cope with the catastrophic forgetting phenomenon. For example, at the beginning of the experiments, the robot is unable to find any meaningful information of an unknown detected object, if it has been never encountered before. After some learning sessions, it has become able to retrieve accurate information about it. 
The learning process of the robot is human-driven, and the human is no more required to be an expert. This allows the application of the system in many dynamic scenarios when a robot needs to learn information about the operating environment.  
Despite the data that drive the learning being sparse and unbalanced, the designed architecture allows the learning curve to converge quickly.
The whole architecture, in addition to these improvements, would make the interactions between humans and robots more natural, making a further step toward the creation of systems that can handle long interactions with humans in an environment whose knowledge of it is incrementally built during the interaction, and it is not needed to give it in advance to the robot.

%
%
%
\bibliographystyle{splncs04}
\bibliography{biblio}

\end{document}